\documentclass[pdflatex,sn-basic,iicol]{sn-jnl}

\usepackage[T1]{fontenc}
\usepackage{graphicx}
\usepackage{amsmath,amssymb,amsfonts}
\usepackage{amsthm}
\usepackage{mathrsfs}
\usepackage{booktabs}
\usepackage{multirow}
\usepackage{xcolor}
\usepackage{algorithm}
\usepackage{algorithmicx}
\usepackage{algpseudocode}
\usepackage{colortbl}
\newcommand{\gcell}[1]{\cellcolor{gray!12}#1}
\usepackage{enumitem}

\usepackage{newtxtext}
\usepackage{newtxmath}
\usepackage[nopatch=footnote]{microtype}
\usepackage{titlesec}

\AtBeginDocument{%
  \hypersetup{
    colorlinks=true,
    citecolor=black,
    linkcolor=black,
    urlcolor=blue
  }%
}

\usepackage{tabularx}
\usepackage{array}
\newcolumntype{Y}{>{\centering\arraybackslash}X}

\titleformat{\section}
  {\normalfont\bfseries\fontsize{12}{14}\selectfont}
  {\thesection}{0.7em}{}

\titleformat{\subsection}
  {\normalfont\bfseries\fontsize{11}{13}\selectfont}
  {\thesubsection}{0.7em}{}

\titleformat{\subsubsection}
  {\normalfont\bfseries\fontsize{10}{12}\selectfont}
  {\thesubsubsection}{0.7em}{}

\titlespacing{\section}
  {0pt}{12pt}{7pt}

\titlespacing{\subsection}
  {0pt}{9pt}{5pt}

\titlespacing{\subsubsection}
  {0pt}{7pt}{4pt}

\usepackage{geometry}
\graphicspath{{figures/}}
\renewcommand{\figurecaptionfont}{%
  \normalfont\fontsize{9}{10}\selectfont
}

\renewcommand{\tablecaptionfont}{%
  \normalfont\fontsize{9}{10}\selectfont
}

\makeatletter
\long\def\@tablecaption#1#2{%
  \setbox\tabcapbox\vbox{%
    \hsize=\linewidth
    \tablecaptionfont
    \centering
    {\bfseries #1}\hskip2mm #2\vphantom{y}\par
  }%
  \box\tabcapbox
}
\makeatother

\begin{document}

\title[VA-Assembly]{%
\parbox{\linewidth}{%
\centering
VA-Assembly: View-Aligned Score Guidance\\
for Training-Free 3D Part Assembly\par
}%
}

\author[1]{\fnm{Yeon-Jae} \sur{Kim}}
\email{yeonjaekim@yonsei.ac.kr}

\author[1]{\fnm{Sangwon} \sur{Kang}}
\email{sangwon.kang@yonsei.ac.kr}

\author*[1]{\fnm{Byung-Kwon} \sur{Min}}
\email{bkmin@yonsei.ac.kr}

\affil[1]{
\orgdiv{School of Mechanical Engineering},
\orgname{Yonsei University},
\orgaddress{
\city{Seoul},
\country{Republic of Korea}%
}%
}

\abstract{
Image-to-3D foundation models have substantially improved single-image 3D reconstruction, but typically produce monolithic representations without explicit part-level structure. Existing part-aware generation and learning-based assembly methods can introduce such structure, but often require part-level supervision, category-specific priors, or paired part-layout training data. We instead consider a practical but underexplored setting in which independently prepared or generated 3D parts have no predefined inter-part relationships, shared coordinate frames, or consistent scales, and must be assembled into a reference-consistent object from a single image. We present VA-Assembly, a training-free framework for 3D part assembly through part-level transformation estimation using view-aligned diffusion guidance. Given a reference image and fixed 3D part geometries, VA-Assembly optimizes each part's translation, rotation, and scale while keeping the geometries unchanged and explicitly separable. The key idea is to condition a frozen view-conditioned diffusion model on same-view renderings of a reference-derived monolithic 3D representation, producing score gradients that are geometrically aligned with the current part configuration. This view-aligned guidance provides multi-view structural cues for transformation optimization under large initial misalignment. By repurposing pretrained visual-structural priors as transformation-level supervision, VA-Assembly requires no annotated part layouts, predefined part taxonomies, category-specific assembly training, or bounding-box-based layout initialization. Experiments compare VA-Assembly with geometric registration and compositional layout baselines using each baseline's supported transformation components, showing improved part-wise transformation estimation. Ablation studies further show that view-aligned conditioning, rather than diffusion guidance alone, is critical for obtaining reliable transformation gradients. Because VA-Assembly updates only part transformations while keeping input geometries fixed, the recovered configurations preserve explicit component separability and support downstream component-level editing, replacement, and reconfiguration.
}


\maketitle

\section{Introduction}
\label{sec:introduction}

Recent 3D foundation models, including diffusion-based view generators, feed-forward reconstruction networks, and large-scale 3D generative models, have substantially improved the quality and accessibility of single-image 3D asset creation \citep{li2025triposg,wang2024crm,xiang2025trellis,lin2023magic3d,long2024wonder3d}. These models make plausible monolithic 3D reconstructions increasingly easy to obtain and often provide coherent estimates of an object's global shape and layout. However, most existing methods still represent the generated object as a single fused geometry or scene representation, without explicit part decomposition or semantic structure \citep{tertikas2023partaware}. Such representations are useful as global object reconstructions, but they obscure the compositional relationships among object parts and limit downstream manipulation. For many applications in graphics, robotics, augmented reality, and computer-aided design, it is desirable for individual components to remain separately replaceable and reconfigurable, rather than being fused into a single inseparable geometry \citep{rodrigues2018part,yao2020front2back,li2020assembly}.

Part-aware 3D generation has emerged as a natural direction to address this issue by representing objects as collections of separable components. These methods typically lift 2D part segmentations into 3D or directly generate part-structured assets using multi-view diffusion, segmentation, or completion networks \citep{kim2024partstad}. While these approaches have shown promising results, they often require specialized part-level supervision, depend on large-scale annotated part datasets, or provide limited control over the arrangement of separately prepared parts \citep{mo2019partnet,li2024pasta}. They leave the spatial arrangement of separately prepared 3D parts underexplored, particularly how such parts can be assembled into a coherent structure consistent with a visual reference.

A part-wise preparation and arrangement strategy offers a practical route toward structured 3D asset creation with explicit component-level control. Components may be manually modeled, retrieved from asset repositories, or generated with off-the-shelf image-to-3D models, and then assembled into a final composition. This workflow preserves component-level editability because each part remains an independent asset. Prior compositional layout and part assembly methods have investigated how independent elements can be arranged into coherent structures. Layout-oriented methods often rely on language-level spatial relations or single-view image-space reconstruction losses, while learning-based assembly methods predict part poses from category-level training data. However, such approaches are not directly suited to reference-consistent intra-object assembly in the full transformation space. Their supervision can be too coarse, view-limited, or dataset-dependent for optimizing fine-grained part contacts, orientations, scales, and inter-part alignments in the full transformation space.

To address this problem, we present VA-Assembly, a training-free framework for reference-guided part-level 3D transformation estimation. VA-Assembly targets a setting in which the input parts are independently prepared or generated and therefore do not share a common coordinate frame, consistent scale, canonical orientation, or predefined inter-part spatial relationship. Unlike layout refinement methods that rely on bounding-box-derived initialization or restricted transformation spaces, VA-Assembly estimates each part's translation, rotation, and scale from a single reference image while keeping the input geometries fixed and explicitly separable.

VA-Assembly uses the reference image to derive view-aligned conditioning signals for transformation optimization. An off-the-shelf image-to-3D model first produces a reference-derived monolithic 3D scaffold, which is rendered from the same camera viewpoints as the current part configuration. These same-view renderings condition a frozen view-conditioned diffusion model, ensuring that each score is evaluated against reference cues expressed in the same rendered view as the optimization variable. This is critical for part transformation estimation: the score must indicate how the currently rendered parts should translate, rotate, and scale, not merely predict a plausible target view from a fixed source image. By aligning the conditioning view with the optimization view, VA-Assembly makes the diffusion score responsive to view-specific discrepancies in silhouette, extent, proportion, and inter-part layout. The gradients are then backpropagated through differentiable rendering to update only the part transformation parameters, yielding multi-view structural guidance under large initial misalignment.

Our training-free formulation for reference-guided part transformation optimization leads to the following contributions:

\begin{itemize}[
    leftmargin=3.7em,
    labelindent=1.1em,
    labelsep=0.5em,
    itemsep=2pt,
    topsep=4pt,
    parsep=0pt
]
\item VA-Assembly formulates single-image part-level 3D transformation estimation for independently prepared or generated parts. The input parts are assumed to have no predefined inter-part layout, shared coordinate frame, or bounding-box-derived initialization, and the method estimates each part's transformation while keeping the part geometries fixed and explicitly separable.
\item View-aligned diffusion guidance provides transformation-level supervision by conditioning a frozen view-conditioned diffusion model on same-view renderings of a reference-derived 3D scaffold. Since the conditioning signal and the current part configuration are rendered from matched viewpoints, the resulting diffusion score is evaluated against reference cues expressed in the same rendered view as the optimization variable, yielding gradients suited for updating part translation, rotation, and scale.
\item The proposed formulation is evaluated against geometric registration and compositional layout baselines using the transformation components supported by each method. Experiments cover translation-rotation, translation-scale, and full translation-rotation-scale settings, while ablation studies show that view-aligned conditioning is more effective than RGB supervision, single-view diffusion guidance, and conventional multi-view diffusion guidance under large initial misalignment.
\end{itemize}

\section{Related Work}
\label{sec:related}

\subsection{Image-to-3D Generation and 3D Foundation Models}

Progress in diffusion-guided 3D generation and multi-view reconstruction has enabled high-fidelity synthesis of textured 3D assets from images, or generated multi-view observations \citep{zhang2022mvgan,hu2024mvdfusion}. Early optimization-based methods distill pretrained 2D diffusion priors into differentiable 3D representations such as neural radiance fields, enabling text- or image-guided 3D synthesis without requiring large-scale paired 3D supervision \citep{poole2022dreamfusion,rombach2022ldm,mildenhall2021nerf,barron2021mipnerf,chen2023fantasia3d}. Subsequent multi-view diffusion models further improve diffusion-guided 3D generation by modeling correlated novel views, generating consistent multi-view observations, or serving as representation-agnostic 3D priors \citep{shi2024mvdream,liu2024syncdreamer,long2024wonder3d}. More recent approaches further improve 3D generation through efficient 3D representations, feed-forward reconstruction networks, and large-scale generative priors trained on 3D or multi-view data \citep{kerbl20233dgs,hong2024lrm,xu2024instantmesh}. Together, these approaches have substantially reduced the effort required to create visually plausible 3D assets across diverse object categories.

View-conditioned image diffusion models provide useful priors for single-image 3D reasoning. Zero-1-to-3 \citep{liu2023zero123} formulates novel-view synthesis as viewpoint-conditioned image translation, synthesizing a target-view image from a source RGB image and a relative camera transformation. However, this source-view conditioning is not designed for reference-consistent part transformation optimization. For target views that are weakly observed or invisible in the source image, Zero-1-to-3 must infer geometry and appearance from its learned image prior, which can introduce hallucinated structure and viewpoint-dependent ambiguity. In multi-view SDS optimization, these independently inferred target-view cues can be inconsistent across sampled views and can provide noisy gradients for part translation, rotation, and scale. VA-Assembly addresses this limitation by conditioning Zero-1-to-3 on target-view renderings of a reference-derived monolithic 3D representation, anchoring each score to reference cues expressed in the same viewpoint as the current rendering.

However, most general-purpose image-to-3D models still represent each object as a single monolithic geometry. The generated asset is typically encoded as a fused mesh, radiance field, Gaussian representation, or volumetric field, without explicit semantic part structure. This makes local editing, part replacement, and compositional reuse difficult, because individual components are not separately represented \citep{tertikas2023partaware,rodrigues2018part}. Moreover, without explicit part semantics, these models provide limited support for reasoning about functional components or reusable submodules \citep{wang2022slotmachine,faruqi2023style2fab,lin2025partcrafter}. These limitations motivate part-aware 3D generation methods, which aim to produce structured assets with explicit component-level organization.

\subsection{Part-Aware 3D Generation Models}

Part-aware 3D generation aims to endow 3D generative models with explicit component-level structure rather than producing a single fused asset \citep{liu2025partfield}. Broadly, existing methods can be grouped into 2D-driven part lifting and 3D-supervised part generation. 2D-driven lifting pipelines first obtain part masks from multi-view images and then lift them into 3D. Some methods invoke 2D foundation models such as the Segment Anything Model (SAM) during reconstruction, while others learn multi-view diffusion modules that jointly segment, complete, and reconstruct parts before reconstruction \citep{ravi2024sam2,chen2025partgen}. Although such methods benefit from strong 2D segmentation and multi-view generation priors, they can suffer from inconsistent part assignments across views, over- or under-segmentation, and limited recovery of occluded or interior geometry. To mitigate cross-view inconsistency, Part123 introduces a contrastively trained part-aware feature field for learning view-consistent part correspondences \citep{liu2024part123}.

In contrast, 3D-supervised part generation methods learn structured shape priors directly in 3D from curated datasets with part-level structure or annotations \citep{mo2019structurenet,koo2023salad,nakayama2023difffacto}. These methods provide stronger part-level control and can generate coherent part-structured shapes within their training distribution. More recent part-aware generation models, such as OmniPart and PartCrafter, further improve semantic decoupling and structural cohesion among generated components \citep{yang2025omnipart,lin2025partcrafter}. However, these methods often rely on category-specific structures, extensive part supervision, or learned part priors, which limits scalability and generalization to open-world compositional objects.

Overall, these methods focus primarily on obtaining part-structured geometry, either by lifting 2D part evidence into 3D or by learning 3D part priors from annotated datasets. Although they do not directly address part transformation optimization, they provide a relevant comparison for evaluating the reference consistency and part-level structure of the final 3D output. We therefore use them as part-aware generation baselines for rendered reconstruction quality and part separability, rather than as direct transformation optimization baselines. VA-Assembly is complementary to this line of work. Instead of learning to synthesize part-structured geometry, it assumes independently prepared part-level assets and focuses on reference-guided arrangement: recovering their translation, rotation, and scale with respect to a monolithic reference 3D structure while preserving part separability.

\subsection{3D Part Alignment and Layout Optimization}

\begin{figure*}[t]
    \centering
    \includegraphics[width=\textwidth]{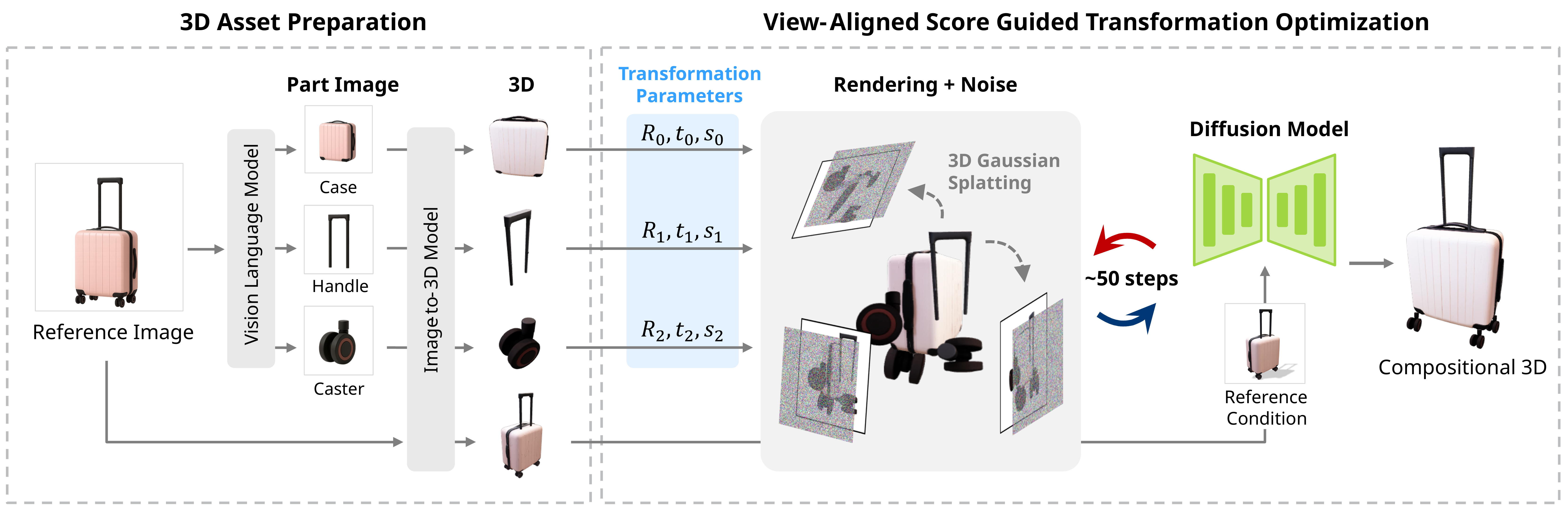}
    \caption{Overview of the VA-Assembly pipeline. 3D assets are generated from a reference image by using a vision-language model and an image-to-3D foundation model. The resulting 3D parts are arranged into a coherent structure via transformation optimization guided by the proposed view-aligned score guidance.}
    \label{fig:overview}
\end{figure*}

3D part alignment is closely related to classical point cloud registration. Methods such as ICP \citep{besl1992registration}, RANSAC-based alignment \citep{fischler1981ransac}, and Go-ICP \citep{yang2015goicp} estimate rigid transformations by establishing geometric correspondences or minimizing distances between point sets. These methods are effective geometry-to-geometry alignment tools when the input point sets have sufficient overlap, clean geometry, and reliable correspondences. Go-ICP further improves robustness by globally optimizing over the rigid motion space, reducing sensitivity to local minima. Certifiable registration methods such as TEASER++ \citep{yang2020teaser} improve robustness to outliers and can estimate similarity transformations, including scale. However, they still rely on reliable geometric correspondences between source and target point sets. In our setting, independently prepared or generated parts may differ from the corresponding region of the monolithic reference 3D in geometry, visible extent, and scale, making correspondence-based registration unstable and sometimes leading to inaccurate scale estimates.

Learning-based assembly methods can predict coherent part poses, but they rely on category-specific training data, large-scale learned assembly priors, or intermediate generative representations. Earlier approaches typically predict part poses directly, often relying on category-specific training data or learned assembly priors \citep{li2020assembly,wu2020pqnet}. Recent methods such as Assembler \citep{zhao2025assembler} improve scalability by formulating part assembly as diffusion-based anchor-point generation, from which part transformations are recovered by least-squares fitting. Although Assembler supports category-agnostic image-conditioned assembly, it remains a learned, data-driven generative model whose final transformations depend on generated anchor points and a subsequent fitting step. In contrast, VA-Assembly is training-free and directly optimizes the transformation parameters of fixed part geometries for each instance using reference-derived view-aligned score guidance.

VA-Assembly also shares conceptual connections with compositional 3D asset generation methods that reconstruct components independently and optimize their spatial layouts. Recent methods such as ComboVerse \citep{chen2024comboverse} and REPARO \citep{han2025reparo} decompose an input image into individual objects, reconstruct each object separately, and then refine their object-level layouts. These methods are effective for compositional asset generation when reliable object-level cues, such as boxes, masks, depth estimates, or initial image-plane layouts, are available. However, their supervision is primarily designed for object-level layout refinement rather than full part-level transformation recovery. ComboVerse relies on 2D layout cues and text-token-attention-based spatial diffusion guidance, while REPARO optimizes translation and scale but excludes rotation in its final layout optimization. These design choices make them less directly applicable to our setting, where independently prepared parts may have arbitrary origins, orientations, scales, and no reliable initial inter-part arrangement.

Overall, existing approaches address only partial aspects of the setting considered in this work. Registration methods depend on reliable 3D correspondences, learning-based assembly methods rely on learned assembly priors or intermediate generative representations, and compositional layout methods often assume useful 2D initialization cues or restricted transformation spaces. VA-Assembly instead targets training-free, reference-guided full part-level transformation estimation for independently prepared or generated 3D parts. The input parts are assumed to have inconsistent coordinate frames, arbitrary origins, unknown scales, and no predefined inter-part spatial relationships. VA-Assembly estimates translation, rotation, and scale from a single reference image without bounding-box-derived initialization, annotated part layouts, or category-specific assembly training, placing it in a broader transformation setting under weaker assumptions.

\section{Main Technical Pipeline}
\label{sec:pipeline}
\subsection{Pipeline Overview}

The overall pipeline of VA-Assembly is illustrated in Fig.~\ref{fig:overview}. Given a reference image $I^{\mathrm{ref}}$, VA-Assembly aims to arrange a set of prepared 3D parts $\mathcal{P}=\{P_i\}_{i=1}^{N}$ into a coherent composition that follows the reference structure while preserving each part as an independent component. Since the parts are prepared separately, they generally lie in different canonical coordinate systems and do not share the relative position, orientation, or scale required for the target object. VA-Assembly resolves this mismatch by optimizing a part-wise transformation for each component without modifying its underlying geometry.

For each part $P_i$, the transformation parameter is defined as $\theta_i=\{R_i,t_i,s_i\}$, where $R_i\in\mathrm{SO}(3)$, $t_i\in\mathbb{R}^{3}$, and $s_i\in\mathbb{R}$ denote rotation, translation, and scale, respectively. The transformed part is written as $\widehat{P}_i(\theta_i)=\{\hat{x}\mid x\in P_i\}$. The full compositional object is represented as a collection of transformed parts, $\hat{\mathcal{O}}(\theta)=\{\widehat{P}_i(\theta_i)\}_{i=1}^{N}$. This representation maintains explicit part-wise structure throughout optimization by parameterizing the object as a set of independently transformed components.

VA-Assembly renders the reference-derived 3D reconstruction and the current part composition from matched viewpoints and uses the resulting paired renderings for view-aligned score guidance, as described in Sect. 4. Visibility-filtered geometric alignment and RGB consistency are used only in the final refinement stage to improve local surface and appearance agreement.

\subsection{3D Asset Preparation}

The required assets consist of a monolithic reference reconstruction and a set of part-level 3D assets. These components may be manually specified, retrieved from an external asset repository, or generated from the input image using pretrained foundation models. In our implementation, a vision-language model first identifies semantic components $\mathcal{C}=\{c_i\}_{i=1}^{N}$ from $I^{\mathrm{ref}}$. Each component $c_i$ is isolated to form a part-level reference image $I_i^{\mathrm{part}}$, which is then processed by an image-to-3D generation model. This process yields the part asset set $\mathcal{P}=\{P_i\}_{i=1}^{N}$.

This preparation stage does not estimate inter-part transformations. The generated parts are therefore treated as fixed, independently prepared assets, and their spatial arrangement is recovered only in the subsequent transformation optimization.

\section{View-Aligned Score Guidance for Part Transformation Optimization}
\label{sec:method}

\subsection{Diffusion Score Guidance and Target-View Conditioning}

\begin{figure}[t]
\centering
\includegraphics[width=\columnwidth]{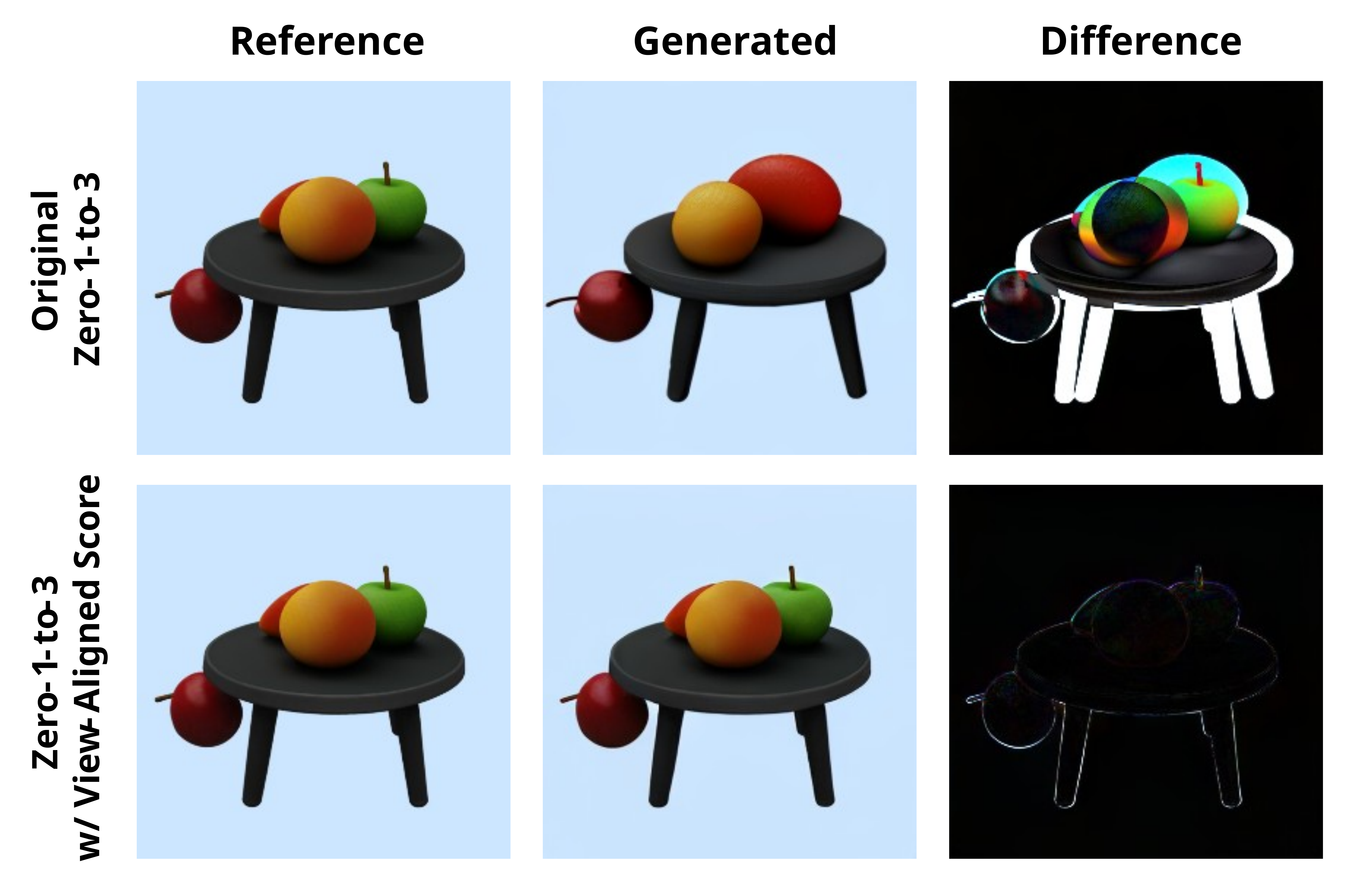}
\caption{Effect of view-aligned score guidance on target-view generation. Compared with original Zero-1-to-3, the proposed guidance better preserves target-view structure. Darker difference maps indicate smaller discrepancies.}
\label{fig:effect_viewscore}
\end{figure}

A central challenge in reference-guided 3D part arrangement is obtaining transformation gradients that are sufficiently global to correct large translation, rotation, and scale errors, while preserving the fixed geometry of each part. Explicit geometric and photometric losses, such as Chamfer distance and RGB reconstruction loss, provide useful constraints when the current arrangement is already close to the reference. However, under large initial misalignment, these losses can be dominated by local visible-surface mismatch or appearance differences and may not provide reliable global directions for correcting the overall layout. We therefore use diffusion score guidance as a structural prior for transformation optimization. The role of SDS in VA-Assembly is not to synthesize new geometry, but to provide score gradients that guide the rendered part composition toward a reference-consistent structure.

Original Zero-1-to-3 conditions on a fixed source image and predicts target-view structure from a relative camera pose. When used directly for multi-view transformation optimization, these target-view predictions can provide inconsistent structural cues across viewpoints, especially in silhouettes, part boundaries, and inter-part spatial relationships. VA-Assembly instead renders the monolithic reference 3D from the same viewpoint as the current part composition and uses this target-view rendering as the conditioning input. This view-aligned conditioning provides viewpoint-specific structural evidence and yields score gradients that are better aligned with the current rendered composition.

The effect of target-view conditioning is illustrated in Fig.~\ref{fig:effect_viewscore}. Zero-1-to-3 guidance can generate plausible novel-view predictions, but the resulting views may not provide sufficiently consistent or viewpoint-sensitive structural signals for transformation optimization. Nearby target poses can yield visually similar predictions, limiting the score variation needed to distinguish small changes in rotation, translation, and scale. By conditioning on target-view renderings of the monolithic reference reconstruction, VA-Assembly supplies viewpoint-specific structural constraints and obtains score guidance that is better aligned with the current rendered composition.

\begin{figure}[t]
    \centering
    \includegraphics[width=\columnwidth]{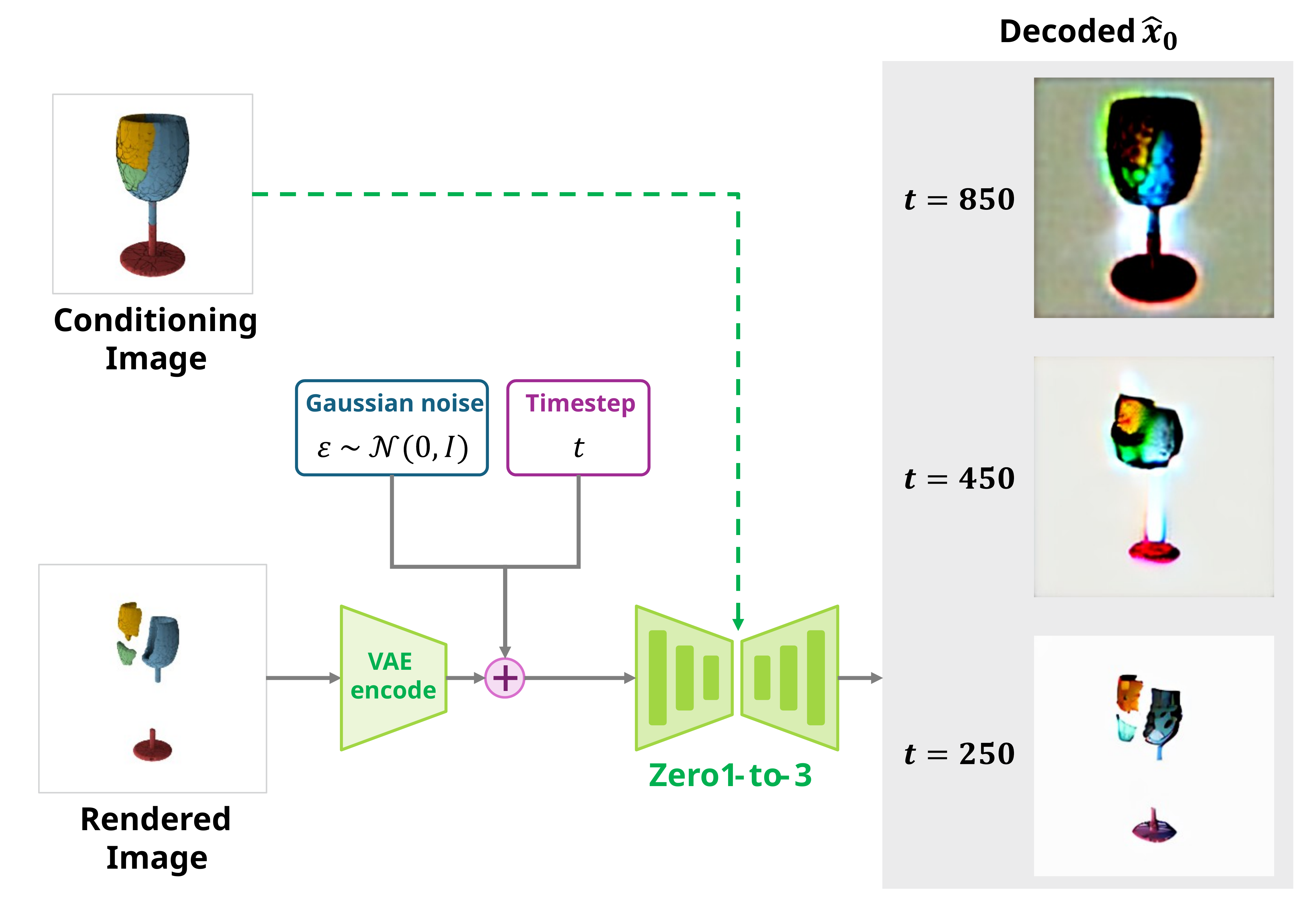}
    \caption{Clean-latent estimates decoded from noisy rendered latents at different diffusion timesteps. High-noise guidance (t=850) yields broad, structure-oriented corrections over the object region, while lower-noise guidance (t=450,250) retains more local details and boundary information.}
    \label{fig:clean_latent_estimates}
\end{figure}

To further clarify why SDS is suitable as the main early-stage guidance, we examine the structural information encoded at different diffusion timesteps. Given the same current rendering and reference condition, we compute the clean-latent estimate ${\hat{x}}_0$ and decode it to image space, as shown in Fig.~\ref{fig:clean_latent_estimates}. Higher-noise scores induce broader layout and silhouette corrections, while lower-noise scores preserve more local boundary information. This behavior supports the use of SDS as early-stage global structural guidance for part arrangement.

\subsection{View-Aligned SDS Objective}

\begin{figure*}[t]
    \centering
    \includegraphics[width=\textwidth]{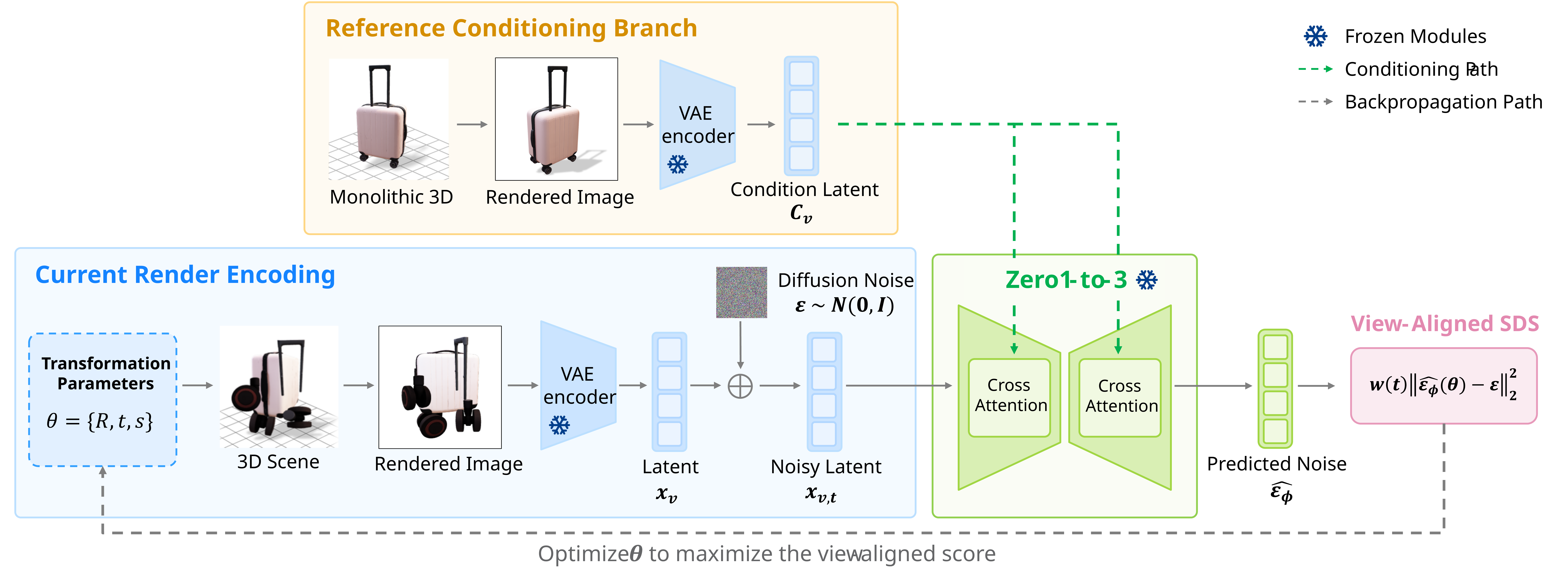}
    \caption{View-aligned score distillation for part transformation optimization. For each sampled view, the reference-view render provides the Zero-1-to-3 conditioning, while the current render is encoded and noised to form $\mathbf{x}_{v,t}$. The frozen diffusion U-Net predicts the noise under this same-view condition, from which the view-aligned SDS loss is computed. Its gradients are backpropagated to the current render to guide part transformation optimization.}
    \label{fig:pipeline}
\end{figure*}

The proposed view-aligned score distillation process is summarized in Fig.~\ref{fig:pipeline}. The current composition rendering is encoded into the diffusion latent space and perturbed according to a sampled noise level. The frozen diffusion model then predicts the noise residual conditioned on the target-view reference, from which the SDS gradient is computed. The resulting score gradients are backpropagated through the differentiable renderer to update only the part transformation parameters.

For viewpoint $v$, the current part composition and the monolithic reference 3D representation are rendered as $I_v(\theta)$ and $I_v^{\mathrm{ref}}$. Both renderings are encoded into the latent space of the diffusion model using the VAE encoder $\mathcal{E}$:

\begin{equation}
x_v(\theta) = \mathcal{E}(I_v(\theta)), \quad C_v = \mathcal{E}(I_v^{\mathrm{ref}})
\end{equation}

In the Zero-1-to-3 conditioning interface, $I_v^{\mathrm{ref}}$ is used as the conditioning image for the same target viewpoint as the current rendering. We therefore set the relative camera transformation condition to the identity view change, so that the diffusion score is evaluated with respect to a same-view reference rather than a novel-view prediction from a fixed source image.

Gaussian noise is then added to the current part-composition latent $x_v(\theta)$ at diffusion timestep $t$ to:

\begin{equation}
x_{v,t}(\theta)
= \alpha_t x_v(\theta) + \sigma_t \epsilon,
\qquad
\epsilon \sim \mathcal{N}(0,\mathbf{I}).
\end{equation}

The diffusion model predicts the noise conditioned on the same-view reference rendering, with the relative camera input fixed to the identity view change:

\begin{equation}
\hat{\epsilon}_{\phi}^{v}
=
\epsilon_{\phi}
\left(
x_{v,t}(\theta), C_v, t
\right)
\end{equation}

We express the view-aligned SDS surrogate objective using the residual between the predicted noise and the injected noise:

\begin{equation}
\mathcal{L}_{\mathrm{VA-SDS}}^{v}(\theta)
=
\mathbb{E}_{t,\epsilon}
\left[
w(t)
\left\|
\hat{\epsilon}_{\phi}^{v}(\theta)
-\epsilon
\right\|_2^2
\right]
\end{equation}

This formulation repurposes Zero-1-to-3 from a standalone novel-view predictor into a reference-anchored score prior for part transformation optimization. The conditioning latent $C_v$ is not a fixed source-view image condition, but a target-view rendering of the monolithic reference 3D. It therefore acts as a same-view geometric anchor for the current rendering $I_v(\theta)$, encouraging score gradients that preserve the reference-view silhouette, part proportions, and inter-part spatial relationships.

The view-aligned SDS gradient can be approximated as

\begin{equation}
\begin{aligned}
\nabla_\theta
\mathcal{L}_{\mathrm{VA-SDS}}^v
\approx {}&
\left(
\frac{\partial I_v(\theta)}
{\partial\theta}
\right)^\top
\left(
\frac{\partial\mathcal{E}(I_v(\theta))}
{\partial I_v(\theta)}
\right)^\top\\
&\quad
w(t)
\left(
\hat{\epsilon}_{\phi}^v(\theta)-\epsilon
\right).
\end{aligned}
\end{equation}

This expression shows that the transformation update is driven by a latent score residual produced by the frozen diffusion model under the target-view reference condition $C_v$. Since the score is evaluated in the VAE latent space under the frozen diffusion prior, it provides high-level structural guidance rather than direct image-space reconstruction supervision. This makes view-aligned SDS effective for early-stage correction of large translation, rotation, and scale errors before local geometric or photometric correspondence becomes reliable.

\subsection{Multi-View Score Aggregation}
SDS gradients are computed in rendered image space and backpropagated through the differentiable Gaussian splatting renderer. In 3DGS, gradients can flow through the projected Gaussian center, projected covariance, depth ordering, and alpha compositing. However, in single-view image-space supervision, the projected-center term often provides the most direct way to change the rendered appearance. Let $\mu_i = (X_i, Y_i, Z_i)$ denote the mean of the $i$-th Gaussian in the camera coordinate system, where $Z_i$ is its depth, and let $(u_i, v_i)$ be its projected image-plane coordinates. With focal length $f$, and omitting the principal point for notational simplicity, the projected center is

\begin{equation}
u_i = f \frac{X_i}{Z_i}, \quad v_i = f \frac{Y_i}{Z_i}
\end{equation}

The relative sensitivity between depth motion and image-plane motion is

\begin{equation}
\frac{\left|\partial u_i / \partial Z_i\right|}{\left|\partial u_i / \partial X_i\right|} = \frac{\left|X_i\right|}{Z_i}, \quad \frac{\left|\partial v_i / \partial Z_i\right|}{\left|\partial v_i / \partial Y_i\right|} = \frac{\left|Y_i\right|}{Z_i}
\end{equation}

\begin{figure}[t]
    \centering
    \includegraphics[width=\columnwidth]{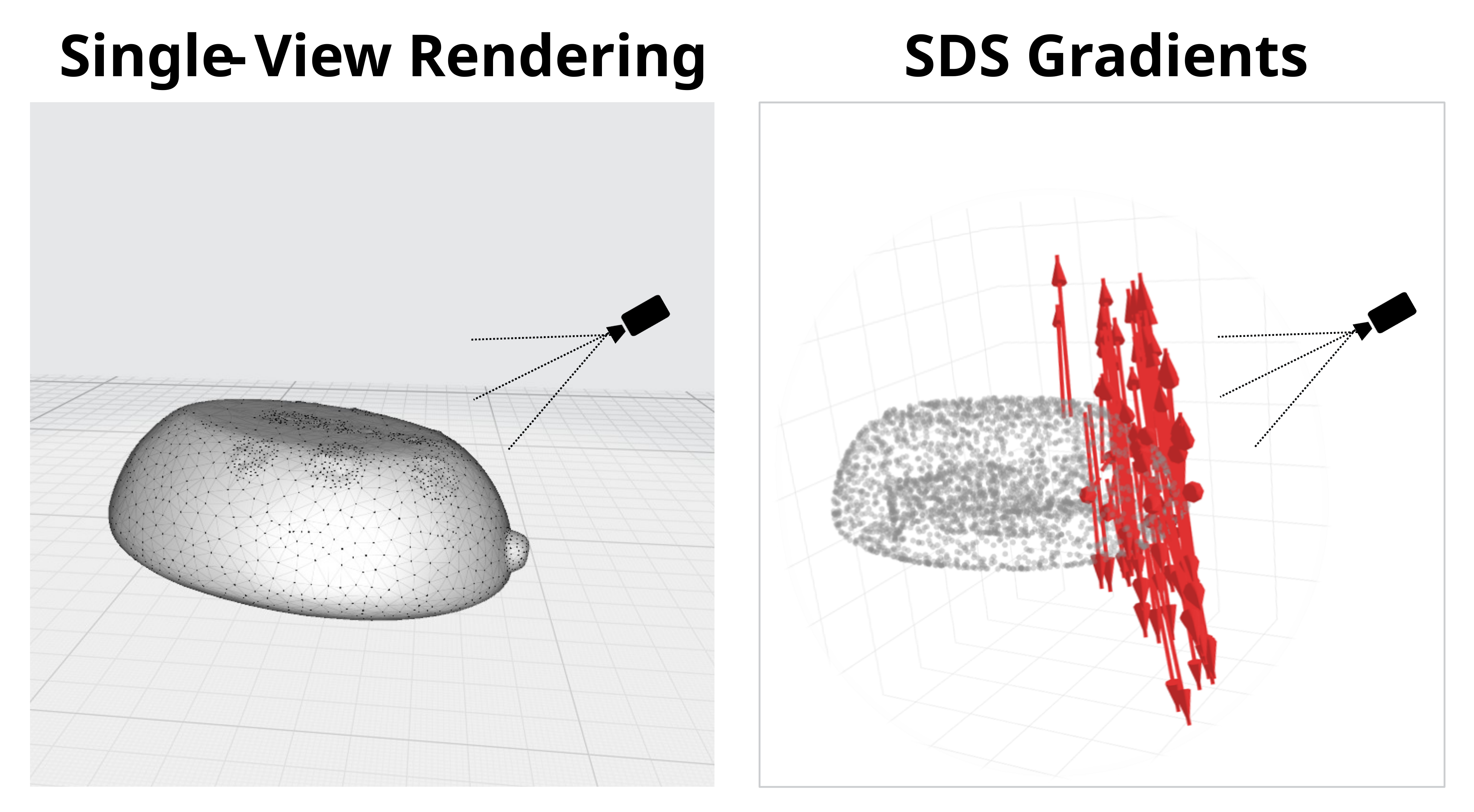}
    \caption{View-Dependent 3D SDS Gradients from Single-View Rendering. Single-view SDS produces projection-biased gradients on 3D points, motivating multi-view aggregation for robust 3D arrangement.}
    \label{fig:sds_gradients}
\end{figure}

These ratios correspond to the normalized image-plane distance from the optical axis. Thus, for Gaussians projected near the image center, depth-direction motion induces weaker projected-center displacement than motion parallel to the image plane. This does not imply that depth gradients are absent, since depth also affects projected covariance, visibility, and alpha compositing. Rather, it explains why single-view SDS gradients can be biased toward screen-space corrections, as visualized in Fig.~\ref{fig:sds_gradients}.

VA-Assembly mitigates this projection-induced bias by aggregating view-aligned score gradients across multiple viewpoints. Each view constrains the visible silhouette and layout from a different projection, so directions weakly constrained in one view can be complemented by gradients from other views. This provides more balanced supervision for optimizing part-wise translation, rotation, and scale in 3D space.

At each optimization iteration, we form a multi-view mini-batch $\mathcal{V}$ by randomly sampling $K = 12$ camera viewpoints from the predefined view distribution. For every sampled viewpoint $v \in \mathcal{V}$, the monolithic reference scaffold and the current part composition are rendered using the same camera, so that each score evaluation is conditioned on a same-view reference rendering. This random multi-view sampling provides complementary silhouette, depth, and occlusion constraints while preventing the optimization from overfitting to a fixed set of projections. The view-aligned SDS objective is averaged over the sampled view mini-batch as

\begin{equation}
\mathcal{L}_{\mathrm{VA-SDS}}(\theta) = \frac{1}{\left|\mathcal{V}\right|} \sum_{v \in \mathcal{V}} \mathcal{L}_{\mathrm{VA-SDS}}^v(\theta)
\end{equation}

\subsection{Late-Stage Geometric and Photometric Refinement}
After the coarse part arrangement is obtained by view-aligned SDS, we apply geometric and photometric losses as late-stage refinement terms. These losses are not intended to drive the global transformation search from a severely misaligned initialization. Instead, they are introduced after SDS-guided optimization has produced a roughly reference-consistent configuration, where local surface correspondence and appearance consistency become more reliable.

\subsubsection{Visibility-Filtered Chamfer Distance}
View-aligned score guidance provides semantic and spatial optimization signals, but it does not explicitly enforce geometric correspondence. We therefore use a visibility-filtered Chamfer distance to align the visible geometry of the transformed parts with the reference 3D. For each viewpoint $v$, Gaussian means are projected onto the image plane and filtered by depth ordering. Let $P_v(\theta) = \{p_j(\theta)\}_{j=1}^{M_v}$ denote the visible Gaussian means of the current part composition and let $Q_v = \{q_k\}_{k=1}^{K_v}$ denote the visible reference points obtained from the monolithic reference 3D. The single-view Chamfer distance is defined as

\begin{equation}
\begin{aligned}
\mathcal{L}_{\mathrm{vCD}}^v(\theta)
={}&
\frac{1}{M_v}
\sum_{j=1}^{M_v}
\min_{q\in Q_v}
\|p_j(\theta)-q\|_2^2\\
&+
\frac{1}{K_v}
\sum_{k=1}^{K_v}
\min_{p\in P_v(\theta)}
\|q_k-p\|_2^2.
\end{aligned}
\end{equation}

and the visibility-filtered Chamfer loss is averaged over the sampled viewpoints:

\begin{equation}
\mathcal{L}_{\mathrm{vCD}}(\theta) = \frac{1}{|\mathcal{V}|} \sum_{v \in \mathcal{V}} \mathcal{L}_{\mathrm{vCD}}^v(\theta)
\end{equation}

By computing the Chamfer distance only over visible samples, this loss avoids penalizing occluded geometry and reduces spurious gradients from regions that are not supported by the rendered reference view.

\subsubsection{RGB Loss}
In addition to surface-level alignment, we impose RGB consistency to preserve the rendered appearance of the part composition with respect to the monolithic reference 3D. For each viewpoint $v$, let $I_v(\theta)$ denote the RGB image rendered from the current part arrangement, and let $I_v^{\mathrm{ref}}$ denote the RGB image rendered from the reference 3D. The RGB loss is computed over valid foreground pixels $S_v$ as

\begin{equation}
\mathcal{L}_{\mathrm{RGB}}^v(\theta) = \frac{1}{|S_v|} \sum_{u \in S_v} \|I_v(\theta, u) - I_v^{\mathrm{ref}}(u)\|_1
\end{equation}

When multiple viewpoints are used, the RGB loss is averaged over the sampled viewpoints:

\begin{equation}
\mathcal{L}_{\mathrm{RGB}}(\theta) = \frac{1}{|\mathcal{V}|} \sum_{v \in \mathcal{V}} \mathcal{L}_{\mathrm{RGB}}^v(\theta)
\end{equation}

This term encourages appearance-level consistency between the optimized part composition and the target-view reference rendering, stabilizing the optimization of part translation, rotation, and scale when score-based guidance alone is ambiguous or view-dependent.

\subsubsection{Staged Objective}
VA-Assembly optimizes part transformations using a staged objective. In the early stage, view-aligned SDS serves as the main optimization signal for coarse global arrangement:

\begin{equation}
\begin{aligned}
\mathcal{L}_{\mathrm{total}}(\theta)
={}&
\lambda_{\mathrm{VA-SDS}}
\mathcal{L}_{\mathrm{VA-SDS}}(\theta)\\
&+
\lambda_{\mathrm{vCD}}
\mathcal{L}_{\mathrm{vCD}}(\theta)\\
&+
\lambda_{\mathrm{RGB}}
\mathcal{L}_{\mathrm{RGB}}(\theta).
\end{aligned}
\end{equation}

During this stage, $\lambda_{\mathrm{VA-SDS}}$ is gradually decreased. Near the end of optimization, the visibility-filtered Chamfer and RGB terms are briefly introduced to refine visible-surface alignment and appearance consistency. We optimize the part transformation parameters $\theta$ with gradient-based updates while keeping the part geometries fixed and separable.

\section{Experiments}
\label{sec:experiments}

\begin{table*}[t]
\centering
\caption{Transformation estimation capabilities of related methods}
\label{tab:transformation_capabilities}
\fontsize{9}{11}\selectfont
\renewcommand{\arraystretch}{1.25}
\setlength{\tabcolsep}{3pt}
\begin{tabularx}{\textwidth}{
@{}
>{\raggedright\arraybackslash}p{120pt}  
>{\raggedright\arraybackslash}p{120pt}  
>{\raggedright\arraybackslash}p{90pt}   
YYY
@{}
}
\toprule
Method & Task Type &
Category Dependency & Translation & Rotation & Scale \\
\midrule
CAP-Net \citep{huang2025capnet} & Object Pose/shape Estimation & Category-specific & $\checkmark$ & $\checkmark$ & $\checkmark$ \\
CRISP \citep{shi2025crisp} & Object Pose/shape Estimation & Category-agnostic & $\checkmark$ & $\checkmark$ & $\times$ \\
TEASER++ \citep{yang2020teaser} & 3D Registration & N/A & $\checkmark$ & $\checkmark$ & $\checkmark$ \\
Go-ICP \citep{yang2015goicp} & 3D Registration & N/A & $\checkmark$ & $\checkmark$ & $\times$ \\
CFPA \citep{zhang2026cfpa} & 3D Arrangement & Category-specific & $\checkmark$ & $\checkmark$ & $\times$ \\
Assembler \citep{zhao2025assembler} & 3D Arrangement & Category-agnostic & $\checkmark$ & $\checkmark$ & $\times$ \\
Comboverse \citep{chen2024comboverse} & 3D Arrangement & Category-agnostic & $\checkmark$ & $\checkmark$ & $\checkmark$ \\
REPARO \citep{han2025reparo} & 3D Arrangement & Category-agnostic & $\checkmark$ & $\times$ & $\checkmark$ \\
\rowcolor{gray!12}
Ours & 3D Arrangement & Category-agnostic & $\checkmark$ & $\checkmark$ & $\checkmark$ \\
\bottomrule
\end{tabularx}
\end{table*}

\subsection{Experimental Settings}
\subsubsection{Datasets}
We evaluate VA-Assembly under two complementary settings. For quantitative part transformation estimation, we construct a synthetic benchmark using 3D assets from the Google Scanned Objects (GSO) dataset \citep{downs2022gso} and the Breaking Bad dataset \citep{sellan2022breakingbad}. For GSO, we create synthetic target compositions by randomly arranging sampled assets and treat the applied transformations as ground truth. For Breaking Bad, we use the provided part assemblies and their corresponding part transformations as ground truth. For each test instance, we randomly sample 2-6 assets from this pool and arrange them into a target composition, resulting in 80 target compositions.

For the part-aware 3D generation comparison, we additionally construct an image-based test set with 20 custom reference images generated using GPT-4o \citep{hurst2024gpt4o}. Each image depicts a multi-part object with visually distinguishable components and is used as the reference input for all compared methods. This setting evaluates the ability to produce reference-consistent 3D objects while preserving explicit part-level structure and part separability.

\subsubsection{Perturbation-based Initialization}
To evaluate robustness under severe initial misalignment, we adopt a deliberately challenging perturbation protocol. After normalizing each complete ground-truth assembly to the $[-1,1]^3$ coordinate cube, we perturb each part independently in translation, rotation, and scale. Specifically, we add Gaussian translation noise with $\sigma_t=0.5$ in normalized coordinates, rotation noise with $\sigma_r=60^\circ$, and log-scale noise with $\sigma_{\log s}=0.4$. This perturbation protocol creates severe initial errors in part position, orientation, and scale, allowing us to evaluate whether VA-Assembly can recover coherent part arrangements from substantially misaligned initial states rather than simply refine near-target initializations.

\subsubsection{Implementation Details}
Each part is reconstructed as a 3D Gaussian asset using TRELLIS \citep{xiang2025trellis}. The view-aligned score guidance is computed with Zero-1-to-3 \citep{liu2023zero123}. We optimize the transformation parameters for 50 iterations using a staged objective. For the first 40 iterations, we use only the view-aligned SDS term to drive coarse global arrangement. The visibility-filtered Chamfer distance and RGB consistency terms are activated only during the final 10 iterations to refine visible-surface alignment and appearance consistency. This schedule prevents the explicit geometric and photometric losses from dominating the early arrangement stage, where large transformation errors require global structural guidance.

\subsubsection{Optimization Settings}
We evaluate the compared methods under three different transformation optimization protocols: translation-rotation, translation-scale, and full translation-rotation-scale optimization. In the translation-rotation protocol, only part translations and rotations are optimized. In the translation-scale protocol, only part translations and scales are optimized. In the full protocol, translations, rotations, and scales are optimized jointly. These protocols are used because the compared baselines adopt different transformation parameterizations and therefore support different degrees of freedom. For a controlled comparison, VA-Assembly is evaluated with the same active transformation components as each corresponding baseline, rather than enabling unsupported degrees of freedom only for our method.

\subsection{Baselines and Metrics}
\subsubsection{Baselines}
Table~\ref{tab:transformation_capabilities} summarizes related methods that estimate geometric transformations across object pose/shape estimation, 3D registration, and 3D arrangement. While these methods are broadly relevant to transformation estimation, they do not define the same output space. Object pose/shape estimation methods estimate object-level pose or size in camera, scene, or reference coordinates, whereas VA-Assembly estimates part-level rotation, scale, and translation within a compositional object. We therefore use Table~\ref{tab:transformation_capabilities} to contextualize the broader transformation-estimation landscape, and select quantitative baselines according to their compatibility with the proposed fixed-part transformation setting. Specifically, the selected baselines must operate on given 3D parts or their point-cloud representations and recover explicit transformation parameters. Since these baselines support different subsets of translation, rotation, and scale, each comparison is conducted within the transformation space supported by the corresponding baseline. VA-Assembly is evaluated with the same active transformation components in each comparison. Under these criteria, we compare VA-Assembly with Go-ICP \citep{yang2015goicp}, TEASER++ \citep{yang2020teaser}, Assembler \citep{zhao2025assembler}, and REPARO \citep{han2025reparo} for part transformation estimation.

Go-ICP \citep{yang2015goicp} and TEASER++ \citep{yang2020teaser} serve as 3D registration baselines. They estimate geometric transformations from 3D observations, but do not use reference-image guidance or explicitly enforce image-consistent part-wise assembly. Assembler \citep{zhao2025assembler} and REPARO \citep{han2025reparo} serve as 3D arrangement baselines. Assembler generates image-conditioned assembled anchor points and recovers rigid part poses through least-squares fitting, whereas REPARO optimizes compositional layouts through differentiable rendering. These methods are closely related to part arrangement, but differ from VA-Assembly in their supported transformation components, supervision signals, and reliance on learned or initialization-dependent layout priors.

Because the baselines support different transformation parameters, we evaluate them under separate optimization protocols. For translation and rotation optimization, we compare with Go-ICP and Assembler. For translation and scale optimization, we compare with REPARO \citep{han2025reparo}. For full translation, rotation, and scale optimization, we compare with TEASER++. In each protocol, VA-Assembly is evaluated using the same active transformation parameters as the corresponding baselines.

We also include quantitative comparisons with part-aware 3D generation methods, OmniPart \citep{yang2025omnipart} and Part123 \citep{liu2024part123}, to evaluate rendered reconstruction quality and part-level structure. Unlike transformation estimation baselines, these methods generate part-structured 3D assets from visual inputs rather than optimizing the transformations of given 3D parts. They are therefore evaluated as part-aware generation baselines, not as direct part transformation optimization baselines.

\subsubsection{Evaluation Metrics}
We use different metrics for the two evaluation settings. For the part alignment benchmark, we evaluate object alignment using Chamfer distance, translation error, rotation error, and scale error. The Chamfer distance is computed between the object point clouds transformed by the optimized and ground-truth transformations, respectively. The translation error ($e_T$) is computed as the Euclidean distance between the predicted and ground-truth translation vectors. For rotation, we report the geodesic angular distance between the ground-truth rotation and the rotation component extracted from the estimated transformation

\begin{equation}
e_R = \cos^{-1}\left(\frac{\mathrm{tr}\left(R_{\mathrm{pred}}^T R_{\mathrm{gt}}\right) - 1}{2}\right)
\end{equation}

reported in degrees. The scale error ($e_S$) is computed as the absolute difference between the predicted and ground-truth scalar scale factors. For all metrics, lower values indicate better performance. We report the mean and standard deviation over all evaluated objects across all test scenes.

For comparison with part-aware 3D generation methods, we evaluate rendered image quality using Peak Signal-to-Noise Ratio (PSNR), Structural Similarity Index Measure (SSIM), and Contrastive Language-Image Pretraining (CLIP) similarity \citep{huynhthu2008psnr, wang2004ssim, radford2021clip}. PSNR measures pixel-wise reconstruction fidelity, SSIM evaluates structural similarity, and CLIP similarity measures semantic consistency in the CLIP embedding space. We additionally provide qualitative comparisons for part separability and editability.

\subsection{Comparison with 3D Part Arrangement Methods}

\begin{figure*}[t]
    \centering
    \includegraphics[width=\textwidth]{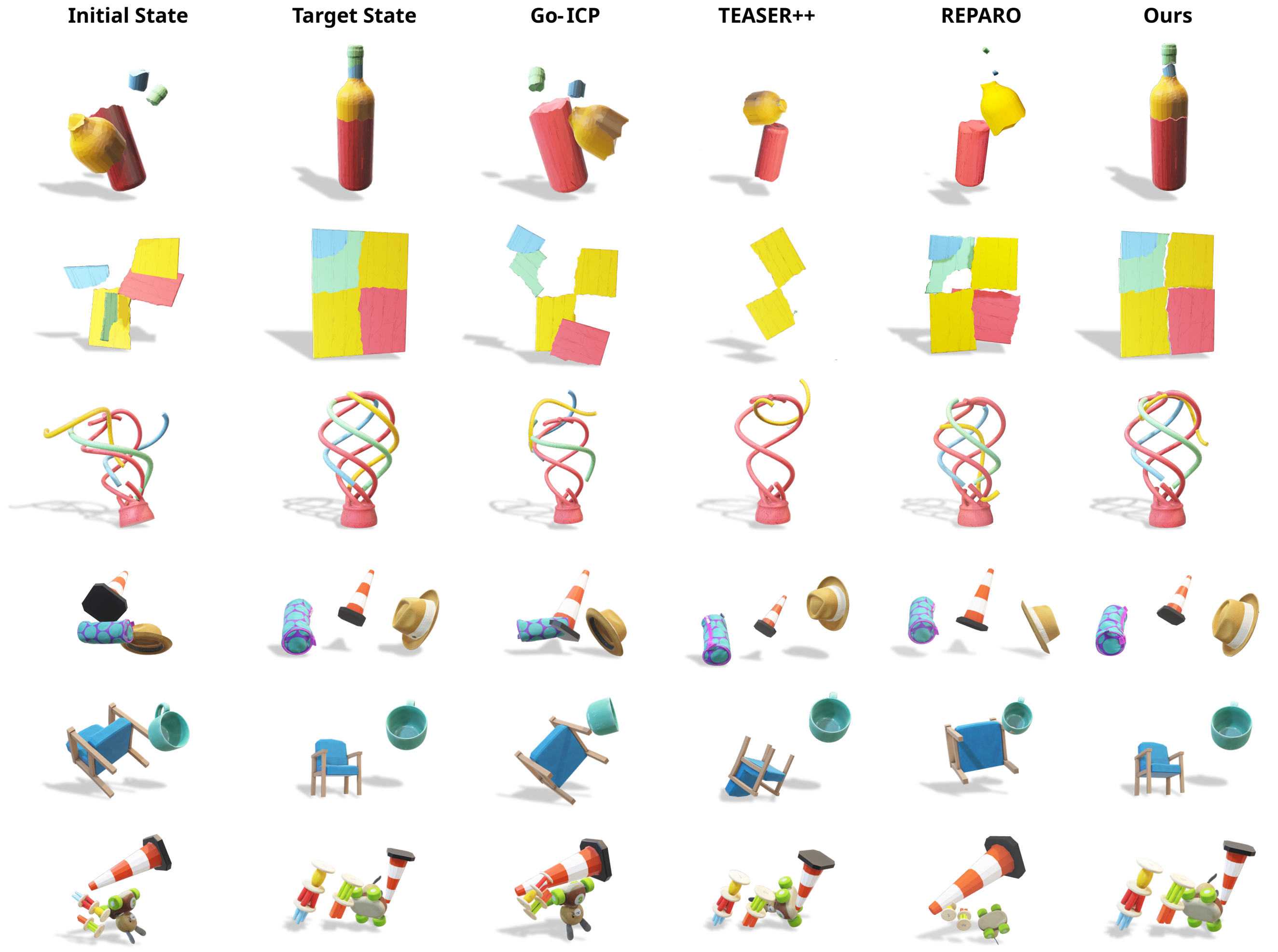}
    \caption{Qualitative comparison on the Breaking Bad and GSO datasets. The top three rows correspond to Breaking Bad, and the bottom three rows correspond to GSO. All methods start from the same challenging initialization, in which the initial 3D part transformations differ from the target in translation, rotation, and scale. VA-Assembly optimizes the full transformation and produces layouts that more closely match the target.}
    \label{fig:qual_comp}
\end{figure*}

\begin{table*}[t]
\centering
\caption{Quantitative comparison within baseline-supported transformation spaces}
\label{tab:quantitative_comparison}
\fontsize{9}{10.5}\selectfont
\setlength{\tabcolsep}{3pt}
\renewcommand{\arraystretch}{1.15}
\begin{tabularx}{\textwidth}
{@{}>{\centering\arraybackslash}m{72pt}c*{8}{Y}@{}}
\toprule
\multirow[c]{2}{*}[-1pt]{\shortstack[c]{Optimization\\parameters}} &
\multirow[c]{2}{*}[-1pt]{Method} &
\multicolumn{2}{c}{Chamf. [NCS] $\downarrow$} &
\multicolumn{2}{c}{Trans. [NCS] $\downarrow$} &
\multicolumn{2}{c}{Rot. [deg] $\downarrow$} &
\multicolumn{2}{c}{Scale $\downarrow$} \\
\cmidrule(lr){3-4}
\cmidrule(lr){5-6}
\cmidrule(lr){7-8}
\cmidrule(l){9-10}
& & Mean & SD & Mean & SD & Mean & SD & Mean & SD \\
\midrule
\multirow{3}{*}{\shortstack{Translation,\\Rotation}}
& Go-ICP
& 0.181 & 0.179 & 0.269 & 0.280
& 44.61 & 28.324 & -- & -- \\
& Assembler
& 0.037 & 0.004 & \textbf{0.049} & 0.001
& 61.038 & 25.156 & -- & -- \\
& \gcell{Ours}
& \gcell{\textbf{0.019}}
& \gcell{0.007}
& \gcell{0.051}
& \gcell{0.013}
& \gcell{\textbf{14.061}}
& \gcell{37.415}
& \gcell{--}
& \gcell{--} \\
\midrule
\multirow{2}{*}{\shortstack{Translation\\Scale}}
& REPARO
& 0.257 & 0.077 & 0.406 & 0.093
& -- & -- & 0.150 & 0.034 \\
& \gcell{Ours}
& \gcell{\textbf{0.017}}
& \gcell{0.009}
& \gcell{\textbf{0.039}}
& \gcell{0.021}
& \gcell{--}
& \gcell{--}
& \gcell{\textbf{0.005}}
& \gcell{0.001} \\
\midrule
\multirow{2}{*}{\shortstack{Translation,\\Rotation, Scale}}
& \mbox{TEASER++}
& 0.063 & 0.027 & 0.074 & 0.063
& 50.843 & 68.169 & \textbf{0.017} & 0.022 \\
& \gcell{Ours}
& \gcell{\textbf{0.021}}
& \gcell{0.007}
& \gcell{\textbf{0.042}}
& \gcell{0.023}
& \gcell{\textbf{18.015}}
& \gcell{42.524}
& \gcell{0.019}
& \gcell{0.020} \\
\bottomrule
\end{tabularx}
\end{table*}

\subsubsection{Quantitative Comparison}
Existing baselines support different transformation components. Go-ICP and Assembler estimate translation and rotation, REPARO optimizes translation and scale, and TEASER++ estimates similarity transformations including translation, rotation, and scale. Therefore, direct comparison in a single full transformation space would either require adding unsupported parameters to some baselines or penalize them for degrees of freedom they do not estimate. To avoid this, we compare each baseline within its supported transformation space and run VA-Assembly with the same active transformation components in the corresponding comparison.

Specifically, we consider three evaluation settings. The translation-rotation setting compares methods that estimate part position and orientation. The translation-scale setting evaluates methods that optimize object extent and placement but do not recover rotation. The full translation-rotation-scale setting evaluates simultaneous optimization of translation, rotation, and scale. This design provides controlled comparisons within each baseline-supported transformation space, while also evaluating VA-Assembly in the full part-level transformation space.

Table~\ref{tab:quantitative_comparison} reports quantitative results where VA-Assembly and each baseline are evaluated using the same active transformation components. VA-Assembly achieves lower object-level Chamfer distance and lower part-level transformation errors in most evaluated settings, indicating that view-aligned diffusion guidance provides effective supervision across different transformation subspaces. The protocols cover both restricted settings, where only a subset of transformation parameters is optimized, and the full setting, where translation, rotation, and scale are estimated jointly.

Compared with geometric registration baselines, VA-Assembly benefits from reference-image guidance rather than relying only on 3D correspondences, which can be ambiguous for independently prepared or generated parts with partial overlap, similar local geometry, or shape discrepancies relative to the reference structure. Compared with arrangement and compositional layout baselines, VA-Assembly directly optimizes the transformations of fixed part geometries and does not rely on generated anchor points, bounding-box-derived initialization, or restricted transformation parameterizations. This allows the method to recover part transformations under larger initial misalignment and weaker assumptions.

The rotation error shows relatively large standard deviations across all methods, mainly due to occasional local-minimum cases involving ambiguous geometry, near-symmetric structures, or weak view-dependent cues. In such cases, parts can be placed in a plausible object-level arrangement while still having incorrect orientations. Despite this variance, VA-Assembly achieves lower mean rotation errors than the baselines, indicating that view-aligned score guidance provides a more reliable signal for part rotation estimation.

\subsubsection{Qualitative Comparison}
Fig.~\ref{fig:qual_comp} presents qualitative results obtained by optimizing the full transformation on the Breaking Bad and GSO datasets. We compare our method with Go-ICP, TEASER++, and REPARO under the same initialization, where the initial part transformations differ from the ground truth in translation, rotation, and scale. Although Go-ICP does not optimize scale and REPARO does not optimize rotation, we include them to assess their behavior under this challenging full-transformation arrangement setting.

VA-Assembly produces layouts that more closely match the target while preserving the overall object structure. In contrast, TEASER++ shows inaccurate scale estimation on the Breaking Bad dataset, likely due to fracture-induced correspondence ambiguity. Specifically, the Fast Point Feature Histograms (FPFH)-based putative correspondences \citep{rusu2009fpfh} provided to TEASER++ can become unreliable around cut surfaces and sharp fracture boundaries, where normals are unstable and local planar regions are geometrically similar. These mismatches contaminate the pairwise distance ratios used for scale voting, reducing the reliability of the global scale consensus.

REPARO can produce plausible layouts in some cases, but it can struggle to recover globally consistent 3D arrangements under our challenging initialization. This limitation is consistent with its single-reference rendering-based supervision, which matches rendered views to a reference image using RGB, predicted depth, image-plane position, and pretrained visual features. Since its final layout optimization updates translation and scale while excluding rotation, REPARO is better suited for refining already plausible object poses than for correcting large discrepancies in rotation, scale, and translation under severe initial misalignment. Similarly, Go-ICP struggles to align parts with their corresponding targets under large initial misalignment, especially when scale correction is required. Overall, these results demonstrate the effectiveness of our view-aligned optimization with geometric constraints for full-transformation 3D part arrangement.

\subsection{Comparison with Part-Aware 3D Generation Methods}

\begin{figure*}[t]
    \centering
    \includegraphics[width=\textwidth]{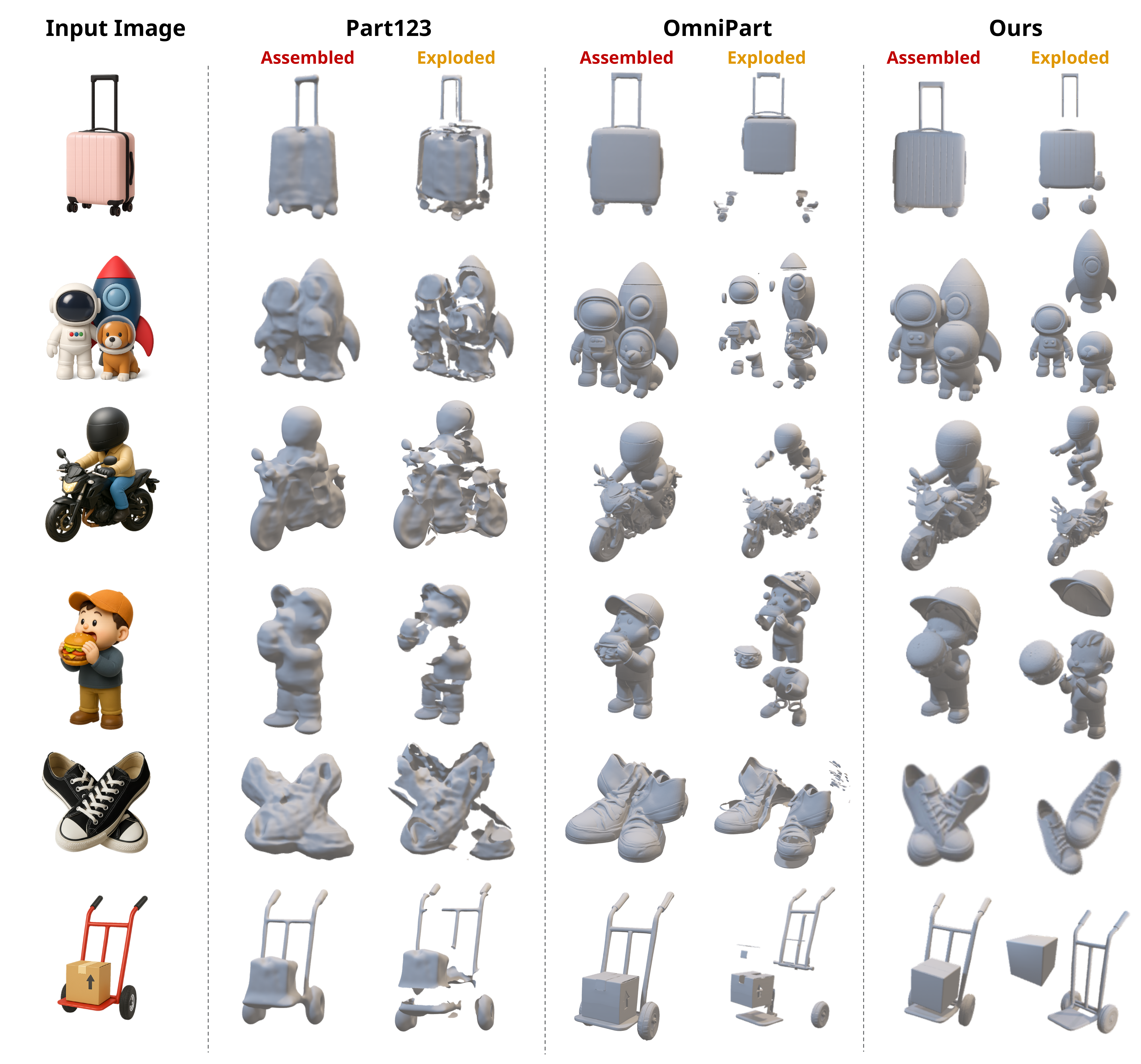}
    \caption{Qualitative comparison with part-aware 3D generation methods. Using the same input image, we compare Part123, OmniPart, and VA-Assembly. Assembled and exploded views are provided to visualize the part-level structure. VA-Assembly preserves explicit part organization and direct part separability.}
    \label{fig:qual_comp_part-aware3D}
\end{figure*}

\begin{table}[t]
\centering
\caption{Quantitative comparison with part-aware 3D generation methods}
\label{tab:generation_comparison}
\fontsize{9}{11}\selectfont
\setlength{\tabcolsep}{3pt}
\renewcommand{\arraystretch}{1.3}
\begin{tabularx}{\columnwidth}
{@{}YYYY@{}}
\toprule
Method & CLIP $\uparrow$ & PSNR $\uparrow$ & SSIM $\uparrow$ \\
\midrule
OmniPart & 0.9021 & \textbf{23.2988} & 0.9172 \\
Part123 & 0.7460 & 18.5328 & 0.9113 \\
\rowcolor{gray!12}
Ours & \textbf{0.9310} & 23.0371 & \textbf{0.9198} \\
\bottomrule
\end{tabularx}
\end{table}

\subsubsection{Quantitative Comparison}
We compare VA-Assembly with OmniPart and Part123, representative part-aware 3D generation approaches. Table~\ref{tab:generation_comparison} reports PSNR, SSIM, and CLIP similarity between rendered views and reference images. VA-Assembly achieves the highest CLIP similarity and SSIM, while obtaining PSNR comparable to OmniPart. These results indicate that VA-Assembly preserves competitive rendered image fidelity and semantic consistency with the reference image. Fig.~\ref{fig:qual_comp_part-aware3D} further shows that this image-level quality is achieved while maintaining explicit part-level organization and editability.

\subsubsection{Qualitative Comparison}
Fig.~\ref{fig:qual_comp_part-aware3D} further shows the difference between direct part-aware generation and our arrangement-based formulation. Part123 tends to generate overly smooth surfaces with limited geometric detail, whereas OmniPart produces visually stronger results but can over-segment objects into ambiguous or fragmented parts. In contrast, our method keeps each semantic component as an independently prepared and arranged 3D asset, enabling direct part-level manipulation such as removal, replacement, and reconfiguration without post-hoc segmentation. Overall, VA-Assembly preserves explicit part organization and direct component-level editability while achieving image-level quality comparable to strong part-aware generation baselines.

\subsection{Ablation Studies}

\begin{table*}[t]
\centering
\caption{Ablation study of RGB supervision and view-aligned score guidance under local and global initialization}
\label{tab:ablation_init}
\fontsize{9}{11}\selectfont
\setlength{\tabcolsep}{8pt}
\renewcommand{\arraystretch}{1.2}
\begin{tabular}{llccc}
\toprule
Init. &
Config. &
Trans. [NCS] $\downarrow$ &
Rot. [deg] $\downarrow$ &
Scale $\downarrow$ \\
\midrule
\multirow{2}{*}{Local}
& $\mathcal{L}_{\mathrm{RGB}}$
& 0.039 & 4.557 & 0.072 \\
& \cellcolor{gray!12}$\mathcal{L}_{\mathrm{VA-SDS}}$
& \cellcolor{gray!12}\textbf{0.035}
& \cellcolor{gray!12}\textbf{3.471}
& \cellcolor{gray!12}\textbf{0.023} \\
\midrule
\multirow{2}{*}{Global}
& $\mathcal{L}_{\mathrm{RGB}}$
& 0.441 & 44.543 & 0.124 \\
& \cellcolor{gray!12}$\mathcal{L}_{\mathrm{VA-SDS}}$
& \cellcolor{gray!12}\textbf{0.053}
& \cellcolor{gray!12}\textbf{13.763}
& \cellcolor{gray!12}\textbf{0.061} \\
\bottomrule
\end{tabular}
\end{table*}

\begin{table*}[t]
\centering
\caption{Ablation study of SDS guidance strategies for part transformation optimization}
\label{tab:ablation_sds}
\fontsize{9}{11}\selectfont
\setlength{\tabcolsep}{5pt}
\renewcommand{\arraystretch}{1.2}
\begin{tabular}{lcccc}
\toprule
SDS Config &
Chamf. $\downarrow$ [NCS] &
Trans. $\downarrow$ [NCS] &
Rot. $\downarrow$ [deg] &
Scale $\downarrow$ \\
\midrule
Single view Zero-1-to-3 & 0.102 & 0.183 & 45.426 & 0.112 \\
Multi view Zero-1-to-3 & 0.111 & 0.234 & 57.881 & 0.078 \\
\rowcolor{gray!12}
Ours: View-aligned SDS & \textbf{0.038} & \textbf{0.057} & \textbf{12.795} & \textbf{0.018} \\
\bottomrule
\end{tabular}
\end{table*}

\begin{figure}[t]
    \centering
    \includegraphics[width=\columnwidth]{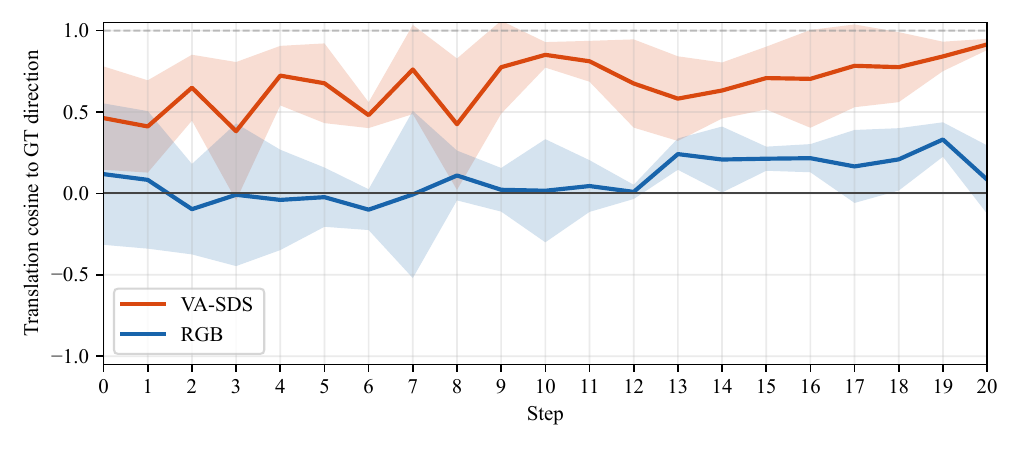}
    \caption{Translation-gradient alignment under global initialization. Cosine similarity is computed between each guidance-induced translation update vector and the ground-truth vector from the current part position to the target position. Higher values indicate better alignment with the correct update direction.}
    \label{fig:translation-gradient}
\end{figure}

We conduct ablation studies to analyze the roles of pixel-level RGB supervision and the proposed view-aligned score guidance under local and global initialization settings. For evaluation, each optimization run is initialized from a controlled random transformation around the target part transformation. In the local initialization setting, we sample initial transformations from a narrow region around the target, with perturbation standard deviations of 0.2 for translation in the normalized object coordinate system, $20^\circ$ for rotation, and 0.2 for log-scale. In the global initialization setting, we sample initial transformations from a wider region around the target, with larger standard deviations of 0.5, $60^\circ$, and 0.5, respectively. These two settings correspond to mild and substantial initial misalignment. For each setting, we report the average translation, rotation, and scale errors in Table~\ref{tab:ablation_init}.

Table~\ref{tab:ablation_init} shows that the two supervision signals play different roles depending on the initialization range. Under local initialization, RGB supervision remains moderately effective because the rendered composition is already close to the target, making pixel-level residuals locally meaningful. However, its performance degrades under global initialization, where large translation, rotation, and scale errors break the image-space correspondence between the current rendering and the reference. In contrast, view-aligned SDS remains effective under both initialization regimes, with a larger advantage under global initialization. Since the score is computed for the current rendering conditioned on target-view reference renderings, it provides image-distribution-level guidance rather than direct pixel-wise residual supervision. This allows the optimization to capture non-local structural discrepancies in silhouette, part layout, and object extent, yielding more reliable update directions under substantial initial misalignment. These results indicate that RGB loss is better suited as a late-stage local refinement term, whereas view-aligned SDS provides the global structural signal required for part transformation optimization.

To further analyze the optimization behavior under global initialization, Fig.~\ref{fig:translation-gradient} compares the translation update directions induced by RGB supervision and view-aligned SDS. At each optimization step, we compute the cosine similarity between each guidance-induced update vector and the ground-truth vector from the current part position to the target position. A value closer to 1 indicates that the update is better aligned with the correct transformation direction. View-aligned SDS consistently shows higher cosine similarity than RGB supervision, indicating that its gradients produce translation updates that are more strongly aligned with the target. This gradient-level analysis supports the quantitative results in Table~\ref{tab:ablation_init} and explains why view-aligned SDS provides more reliable global correction than pixel-level RGB supervision.

We further isolate the effect of the proposed view-aligned score guidance by evaluating SDS-only optimization under three guidance strategies. As shown in Table~\ref{tab:ablation_sds}, we compare single-view Zero-1-to-3, multi-view Zero-1-to-3, and our View-aligned SDS. All variants use pretrained Zero-1-to-3 to estimate the SDS score, but differ in how the guidance signal is conditioned. In the single-view Zero-1-to-3 setting, optimization is performed using only the canonical-view rendering, where both azimuth and elevation are set to $0^\circ$. In the multi-view Zero-1-to-3 setting, the conditioning image is fixed, while target views are specified by randomly sampled relative camera poses. Pretrained Zero-1-to-3 then provides pose-conditioned SDS guidance for the sampled views based on the fixed input image. In contrast, our View-aligned SDS conditions pretrained Zero-1-to-3 on target-view renderings of the reference-derived monolithic reconstruction, so that each score evaluation uses a reference rendering aligned with the sampled viewpoint. Since all conditioning images are rendered from the same monolithic reconstruction, they provide consistent cues about object appearance, silhouette, and spatial layout across views. This leads to more coherent multi-view guidance for transformation optimization. To focus on the score signal itself, this experiment excludes the refinement losses used in the full optimization stage, including the visibility-filtered Chamfer and RGB losses.

Table~\ref{tab:ablation_sds} isolates the effect of view-aligned conditioning. The single-view and conventional multi-view Zero-1-to-3 variants use the same pretrained diffusion prior, but condition the score on either a fixed canonical view or a fixed source image with relative target-view cameras. These variants can provide plausible view-dependent guidance, but the resulting score is not necessarily evaluated against reference cues expressed in the same rendered view as the current optimization variable. As a result, the score may encourage visually plausible target views without providing reliable directions for updating the current part translation, rotation, and scale.

By contrast, view-aligned SDS conditions each score evaluation on a reference rendering from the same camera viewpoint as the current part configuration. This makes the score residual sensitive to view-specific discrepancies in silhouette, object extent, part proportions, and inter-part layout. The large performance gap in Table~\ref{tab:ablation_sds} therefore shows that the improvement does not come from using diffusion guidance alone, but from aligning the conditioning view with the optimization view. This directly supports the central design of VA-Assembly as a transformation-level use of a pretrained view-conditioned diffusion prior.

\subsection{Extensions}
VA-Assembly can be extended to hierarchical assembly for complex multi-part objects. Rather than optimizing all parts against a single product-level reference image, the object can be assembled through multiple stages using references at different hierarchy levels. Since VA-Assembly preserves each optimized component as an independent 3D asset, several lower-level parts can first be arranged into an intermediate sub-assembly and then treated as a higher-level component in subsequent optimization stages.

\begin{figure}[t]
    \centering
    \includegraphics[width=\columnwidth]{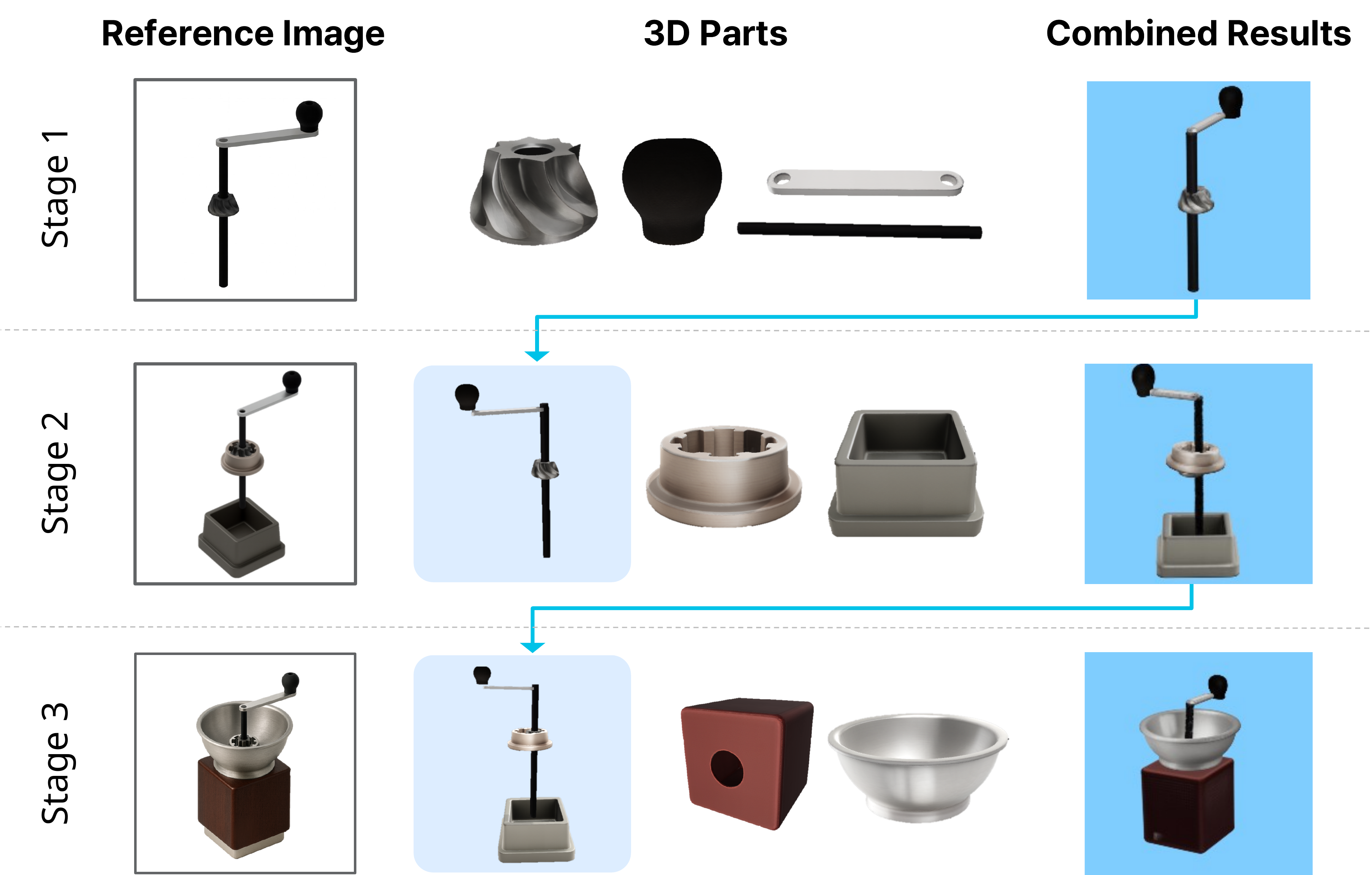}
    \caption{Hierarchical 3D Part Assembly for Product Design.}
    \label{fig:extensions}
\end{figure}

This hierarchical strategy is useful when a single exterior reference image does not provide sufficient information about the placement of internal or occluded components. The exterior view may specify the overall object appearance, but it may not reveal the spatial arrangement of hidden parts. To address this, sub-assembly references can be provided for structures that are not fully observable in the product-level reference. VA-Assembly first optimizes these parts with respect to their corresponding sub-assembly reference and then groups the optimized result as a composite component for higher-level assembly, while preserving the relative transformations of its constituent parts.

Fig.~\ref{fig:extensions} shows an example of this hierarchical workflow for a manual coffee grinder. Internal or partially exposed parts are first optimized using a sub-assembly reference, and the resulting sub-assembly is represented as a composite component with its own group-level transformation. This composite component is then optimized together with other higher-level parts against an exterior product-level reference image. In this way, VA-Assembly supports modular assembly in which internal substructures and exterior components are optimized at different hierarchy levels while remaining independently accessible and controllable.

\section{Conclusion}
\label{sec:conclusion}

This paper investigated training-free, reference-guided 3D part assembly from a single image, formulated as part-level transformation estimation for fixed, separable 3D parts. VA-Assembly addresses the setting in which independently prepared or generated parts have no shared coordinate frame, consistent scale, canonical orientation, predefined inter-part layout, or bounding-box-derived initialization. By using view-aligned diffusion guidance, the framework estimates each part's translation, rotation, and scale while keeping the input geometries fixed and explicitly separable. The experiments show that this formulation improves part-level transformation estimation over geometric registration and compositional layout baselines under transformation-specific evaluation protocols. The ablation studies further clarify the source of this improvement: RGB supervision mainly provides local refinement, whereas view-aligned SDS supplies a global structural signal under large initial misalignment; moreover, the performance gap over single-view and conventional multi-view Zero-1-to-3 guidance shows that the key factor is not diffusion guidance alone, but aligning the conditioning view with the rendered optimization view. These findings suggest that view-aligned diffusion priors can be repurposed as transformation-level supervision for fixed, separable 3D parts. The recovered transformations also support downstream component-level editing, replacement, and reconfiguration. Future work will extend this transformation-centric formulation toward stronger contact reasoning, collision handling, and physically constrained multi-part assembly.

\backmatter

\section*{Declarations}

\noindent
\textbf{Funding}\par
\smallskip
\noindent
This work was partially supported by the Robot Industry Technology Development Program (No. RS-2024-00507501), funded by the Ministry of Trade, Industry and Resources (MOTIR, Korea), and by the National Research Foundation of Korea (NRF) grant funded by the Korea government (MSIT) (No. RS-2023-00222166).

\medskip

\noindent
\textbf{Data Availability}\par
\smallskip
\noindent
The Google Scanned Objects dataset (Downs et al., 2022) used in this study is publicly available at \url{https://app.gazebosim.org/GoogleResearch}. The Breaking Bad dataset (Sellán et al., 2022) is publicly available at \url{https://breaking-bad-dataset.github.io/}. The custom GPT-4o-generated reference images used for image-conditioned evaluation were generated during this study and are available from the corresponding author upon reasonable request. 

\bibliography{sn-bibliography}
\end{document}